\documentclass[runningheads]{llncs}
\usepackage{graphicx}
\usepackage{siunitx}
\usepackage{amsmath,amssymb} 
\usepackage{color}
\usepackage{subfig}
\usepackage{graphicx}
\usepackage{subfig}
\newsavebox{\measurebox}
\usepackage{multirow}
\begin{document}

\title{SPNet: Deep 3D Object Classification and Retrieval using Stereographic Projection} 
\titlerunning{SPNet} 

\author{Mohsen Yavartanoo\inst{1}\orcidID{0000-0002-0109-1202} \and
Eu Young Kim\inst{1}\orcidID{ 0000-0003-0528-6557} \and
Kyoung Mu Lee\inst{1}\orcidID{0000-0001-7210-1036}}
%

\authorrunning{M. Yavartanoo et al.} 


\institute{Department of ECE, ASRI, Seoul National University, Seoul, Korea
\url{https://cv.snu.ac.kr/} \\
\email{\{myavartanoo, shreka116, kyoungmu\}@snu.ac.kr}}


\maketitle

\begin{abstract}
We propose an efficient Stereographic Projection Neural Network (SPNet) for learning representations of 3D objects. We first transform a 3D input volume into a 2D planar image using stereographic projection. We then present a shallow 2D convolutional neural network (CNN) to estimate the object category followed by view ensemble, which combines the responses from multiple views of the object to further enhance the predictions. Specifically, the proposed approach consists of four stages: (1) Stereographic projection of a 3D object, (2) view-specific feature learning, (3) view selection and (4) view ensemble. The proposed approach performs comparably to the state-of-the-art methods while having substantially lower GPU memory as well as network parameters. Despite its lightness, the experiments on 3D object classification and shape retrievals demonstrate the high performance of the proposed method.


\end{abstract}

\section{Introduction}
In recent years, success of deep learning methods, in particular, convolutional neural network (CNN), has urged rapid development in various computer vision applications such as image classification, object detection, and super-resolution. Along with the drastic advances in 2D computer vision, understanding 3D shapes and environment have also attracted great attention.

Many traditional CNNs on 3D data simply extend the 2D convolutional operations to 3D, for example, the work of Wu et al. \cite{7298801} which extends 2D deep belief network to 3D deep belief network, or the works of Maturana et al. \cite{7353481} and Sedaghat et al.  \cite{Sedaghat2016OrientationboostedVN} where they extend 2D convolutional kernels to 3D convolutional kernels. Furthermore, Brock et al. \cite{Brock2016GenerativeAD} and Wu \cite{Wu:2016:LPL:3157096.3157106} proposed to build deeper 3D CNNs following the structures from inception-module, residual connections, and Generative Adversarial Network (GAN) to improve the generalization capability. However, these methods are based on 3D convolutions, thereby having high computational complexity and GPU memory consumption.

An alternate approach is based on projected 2D views of the 3D object to exploit established 2D CNN architectures. MVCNN  \cite{7410471} renders multiple 2D views of a 3D object and use them as an input to 2D CNNs. Some other works  \cite{7273863,Sfikas2017ExploitingTP,SFIKAS2018208} propose to use the 2D panoramic views of a 3D shape. However, these methods can only observe partial parts of the 3D object, failing to cover full 3D surfaces.

To address all these limitations, we introduce a novel 3D shape representation technique using stereographic mapping to project the full surfaces of a 3D object onto a 2D planar image. This 2D stereographic image becomes an input to our proposed shallow 2D CNN, thereby reducing substantial amount of network parameters and GPU memory consumption compared to the state-of-the-art 3D convolution-based methods, while achieving high accuracy.

By taking advantage of multiple projected views generated from a single 3D shape, we propose {\em view ensemble} to combine predictions of most discriminative views, which are sampled by our view selection network. On the contrary, Conventional methods \cite{7410471,7273863,Wang2017DominantSC,7780978,Wang2017OCNNOC,SFIKAS2018208} simply aggregate the responses of all multiple views via max or average pooling.


\section{Related Work}
In this section, we review recent deep learning methods for 3D feature learning. These methods are categorized in term of different feature representations; (1) point cloud-based representations, (2) 3D model-based representations, and (3) 2D and 2.5D image-based representations.

\noindent
\textbf{Point cloud-based methods:} While previous works often combine hand-crafted features or descriptors with a machine learning classifier \cite{Golovinskiy:2009:SRO,Teichman2011Towards3O,Behley2012PerformanceOH,Frome}, the point cloud-based methods operate directly on point clouds in an end-to-end manner. In \cite{8099499,qi2017pointnetplusplus,8237361}, the authors  designed novel neural network architectures suitable for handling unordered point sets in 3D.
Features based on point clouds often require spatial neighborhood queries, which can be hard to deal for inputs with large numbers of points.

\noindent
\textbf{3D model-based methods:} Voxel-based methods learn 3D features from voxels which represent 3D shape by the distribution of corresponding binary variables.

In 3D shapeNet  \cite{7298801}, the authors proposed a method which learns global features from voxelized 3D shapes based on the 3D convolutional restricted Boltzmann machine. Similarly, Maturana and Scherer  \cite{7353481} proposed VoxNet which integrates a volumetric occupancy grid representation with a supervised 3D CNN. In a follow-up, Sedaghat et al. \cite{Sedaghat2016OrientationboostedVN} extended VoxNet by introducing auxiliary task. They proposed to add orientation loss in addition to the general classification loss, in which the architecture predicts both the pose and class of the object. Furuya et al.  \cite{Furuya2016DeepAO} proposed Deep Local feature Aggregation Network (DLAN) which combines rotation-invariant 3D local features and their aggregation in a single architecture.

Sharma et al.  \cite{Sharma2016VConvDAEDV} proposed a fully convolutional denoising auto-encoder to perform unsupervised global feature learning. In addition, 3D variational auto-encoders and generative adversarial networks have been adopted by Brock et al.  \cite{Brock2016GenerativeAD} and Wu et al.  \cite{Wu:2016:LPL:3157096.3157106}, respectively. Furthermore, recent works\cite{Wang2017OCNNOC,8100184} exploit the sparsity of 3D input using the octree data structure to reduce the computational complexity and speed up the learning of global features.

\noindent
\textbf{2D/2.5D image-based methods:} Image-based methods have been considered as one of the fundamental approaches in 3D object classification. Light Field descriptor (LFD)\cite{doi:10.1111/1467-8659.00669} by Chen et al. used multiple views around a 3D shape, and evaluates the dissimilarity between two shapes by comparing the corresponding two view sets in a greedy way instead of learning global features by combining multi-view information. Bai et al. \cite{7780912} used a similar approach but using the Hausdorff distance between the corresponding view sets to measure the similarity between two 3D shapes.

Su et al. \cite{7410471} proposed a CNN architecture that aggregates information from multiple views rendered from a 3D object which achieves higher recognition performance compared to single view based architectures.
By decomposing each view sequence into a set of view pairs, Johns et al. \cite{7780783} classified each pair independently and learned an object classifier by weighting the contribution of each pair, which allows 3D shape recognition over arbitrary camera viewpoint.
To perform pooling more efficiently, Wang et al. \cite{Wang2017DominantSC} proposed a dominant set clustering technique where pooling is performed in each cluster individually.
Kanezaki et al. \cite{kanezaki2018_rotationnet} proposed RotationNet which takes multi-view images of an object and jointly estimates its object category and poses. RotationNet learns viewpoint labels in an unsupervised manner. Moreover, it learns view-specific feature representations shared across classes to boost the performance.

As an alternative approach, Gomez-Donoso et al. \cite{7965883} proposed LonchaNet which uses three orthogonal slices from 3D point cloud as an input to three independent GoogLeNet networks, each network learning specific features for each slice. Cohen et al. \cite{s.2018spherical} in Spherical CNNs proposed a definition for the spherical cross-correlation that is both expressive and rotation-equivariant. The spherical correlation satisfies a generalized Fourier theorem, which allows to compute it efficiently using a generalized Fast Fourier Transform (FFT) algorithm. Papadakis et al. \cite{Papadakis2010} proposed PANORAMA that uses a set of panoramic views of a 3D object which describe the position and orientation of the object's surface in 3D space. 2D Discrete Fourier Transform and the 2D Discrete Wavelet Transform are computed for each view. Shi et al. in DeepPano \cite{7273863}, projected each 3D shape into a panoramic view around its principal axis and used a CNN for learning the representations from these views. To make the learned representations invariant to the rotation around the principal axis a row-wise max-pooling layer is applied between the convolution and fully-connected layers.
to achieve better feature descriptor for a 3D object in the training phase, Sfikas et al. \cite{SFIKAS2018208} use three panoramic views corresponding to the major axes and taking average pooling over feature descriptor of each view for the training of an ensemble of CNNs.

\section{Proposed Stereographic Projection Network}
In this section, we provide details of our proposed approach. We first describe how to transform a 3D object into a 2D planar image using stereographic projection. Then, we give the detailed description of the proposed shallow 2D CNN architecture, SPNet, followed by the procedures for view selection and view ensemble.

\subsection{Streographic Representation}
Stereographic projection is a mapping that projects a 2D manifold onto a 2D plane. Such a technique is well developed in the field of Topology and Geography to project surface of the earth to a 2D planar map \cite{nla.cat-vn2145201}. Since then, various projection functions have been proposed to improve the quality of mapping. In this work, we explore different types of projection functions showing that stereographic projection preserves the more detailed surface structure of a 3D object.

To construct the stereographic representation of a 3D object, we first normalize the 3D object such that a unit sphere can fully cover it. We then translate the origin of the sphere to the center of the object assuming that the orientation of the object is aligned. For each point $\mathit{p}$ on the surface of the object, we denote $\mathit{e}$ as a unit vector from the origin $\mathit{o}$ to the point $\mathit{p}$ as shown in Fig.~\ref{fig:streo}(a). By assuming that the poles are aligned with the z-axis, image coordinates in 2D mapped image can be determined by different types of projection functions as follow:

\begin{figure}
\centering
\sbox{\measurebox}{%
  \begin{minipage}[b]{.27\textwidth}
  \subfloat[Chair]{
    {\label{fig:figA}\includegraphics[width=\textwidth,height=3.9cm]{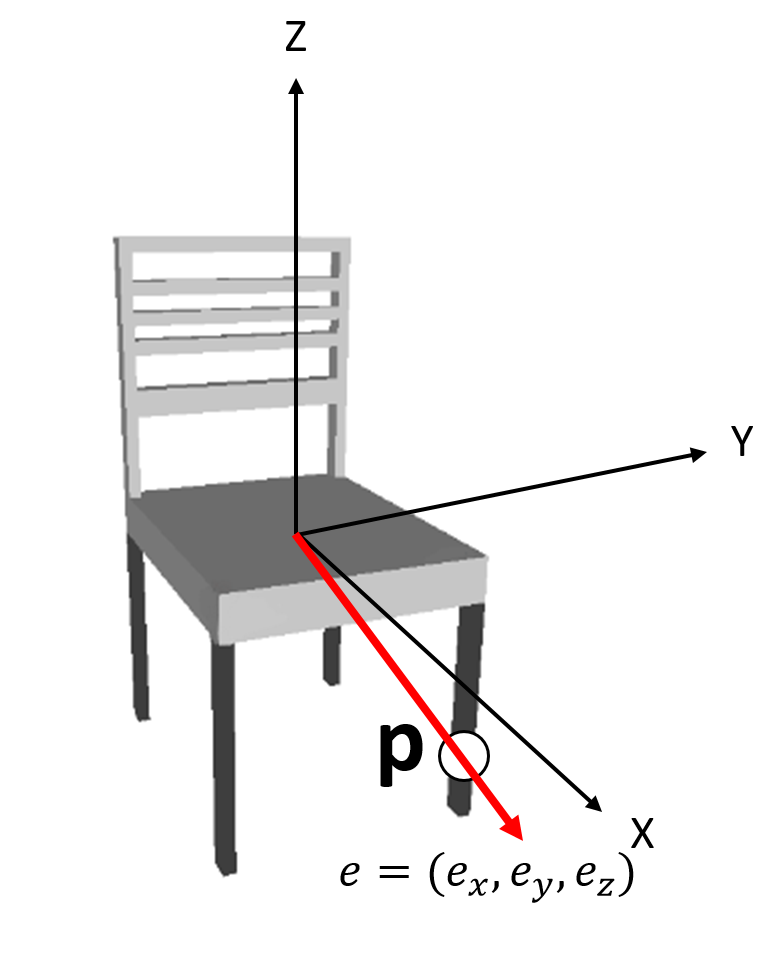}}}
\end{minipage}}
\usebox{\measurebox}\qquad
\begin{minipage}[b][\ht\measurebox][s]{0.15\textwidth}
\centering
  {\label{fig:fig1}\includegraphics[width=\textwidth,height=1.5cm]{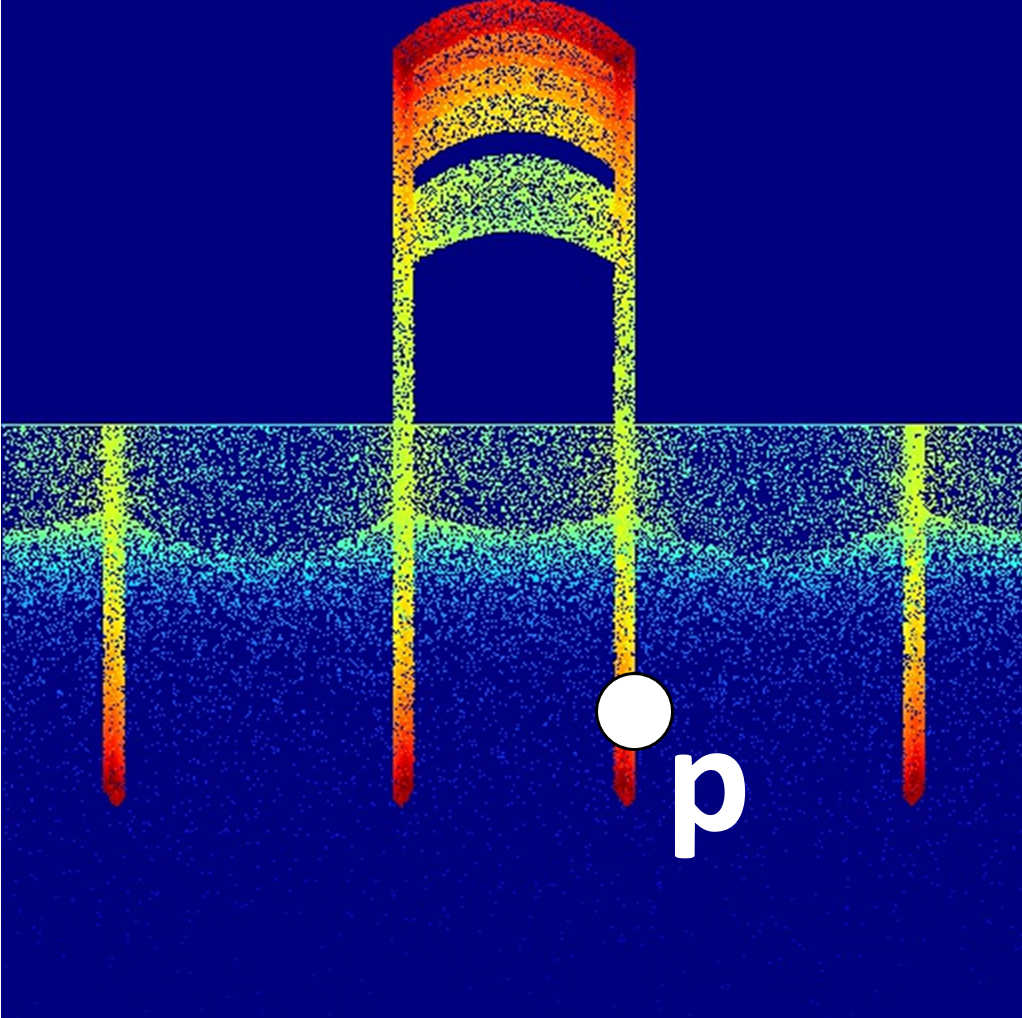}{\caption*{(b)\tiny UV-mapping}}}
\vfill
  {\label{fig:fig2}\includegraphics[width=4cm,height=1.5cm]{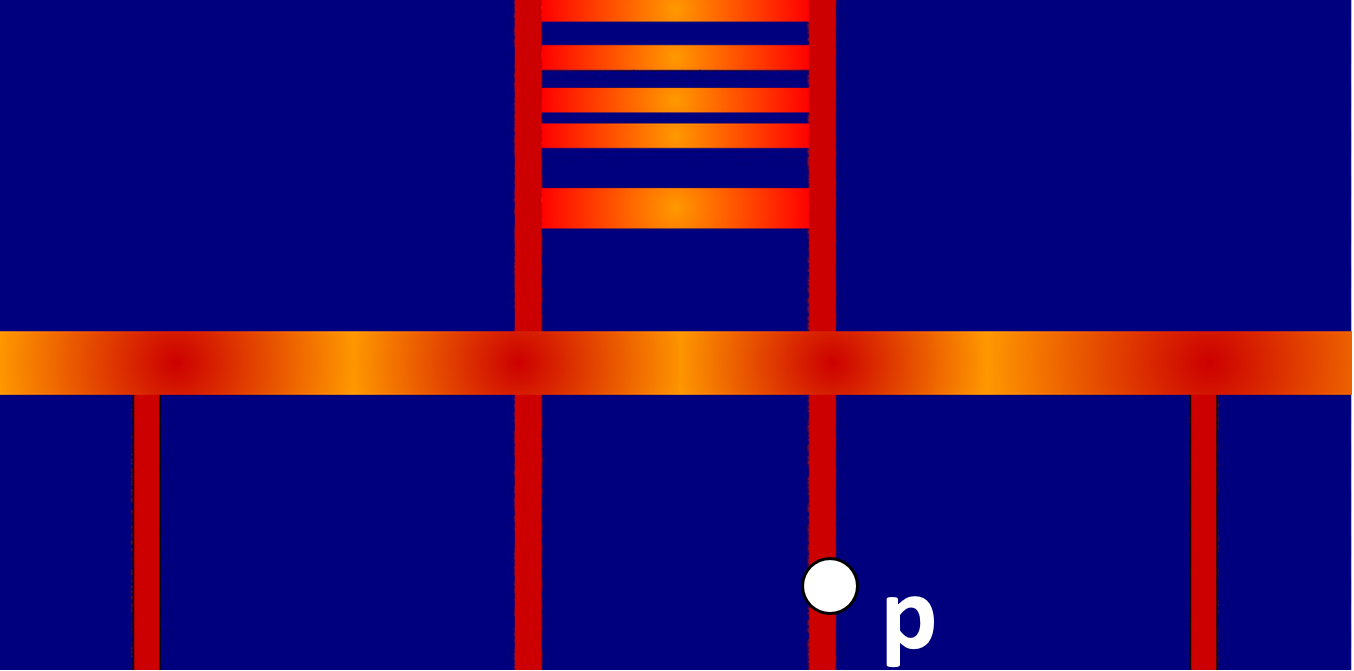}{\caption*{(f)\tiny Panorama}}}
\end{minipage}
\begin{minipage}[b][\ht\measurebox][s]{.17\textwidth}
\centering
  {\label{fig:fig3}\includegraphics[width=\textwidth,height=1.5cm]{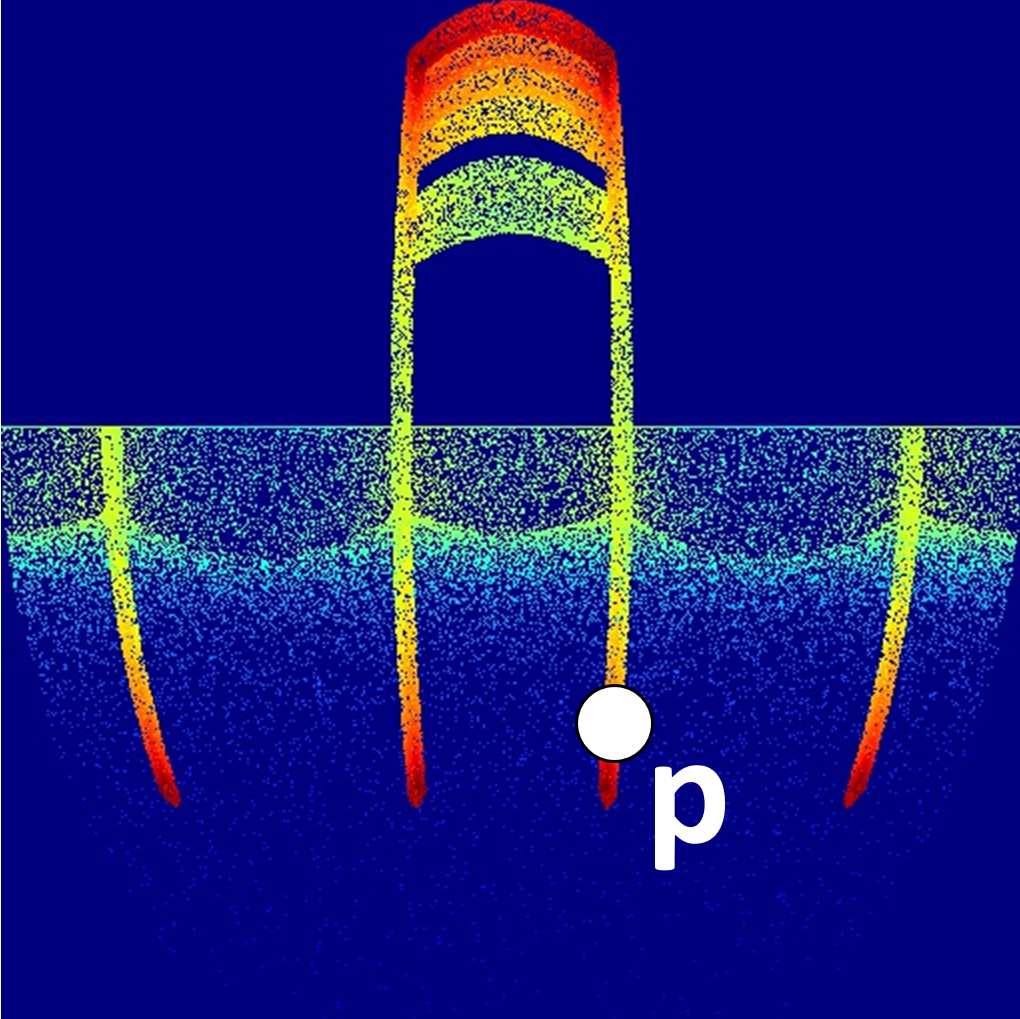}{\caption*{(c)\tiny Kavrayskiy VII}}}
\vfill
\end{minipage}
\begin{minipage}[b][\ht\measurebox][s]{0.15\textwidth}
\centering
  {\label{fig:fig4}\includegraphics[width=\textwidth,height=1.5cm]{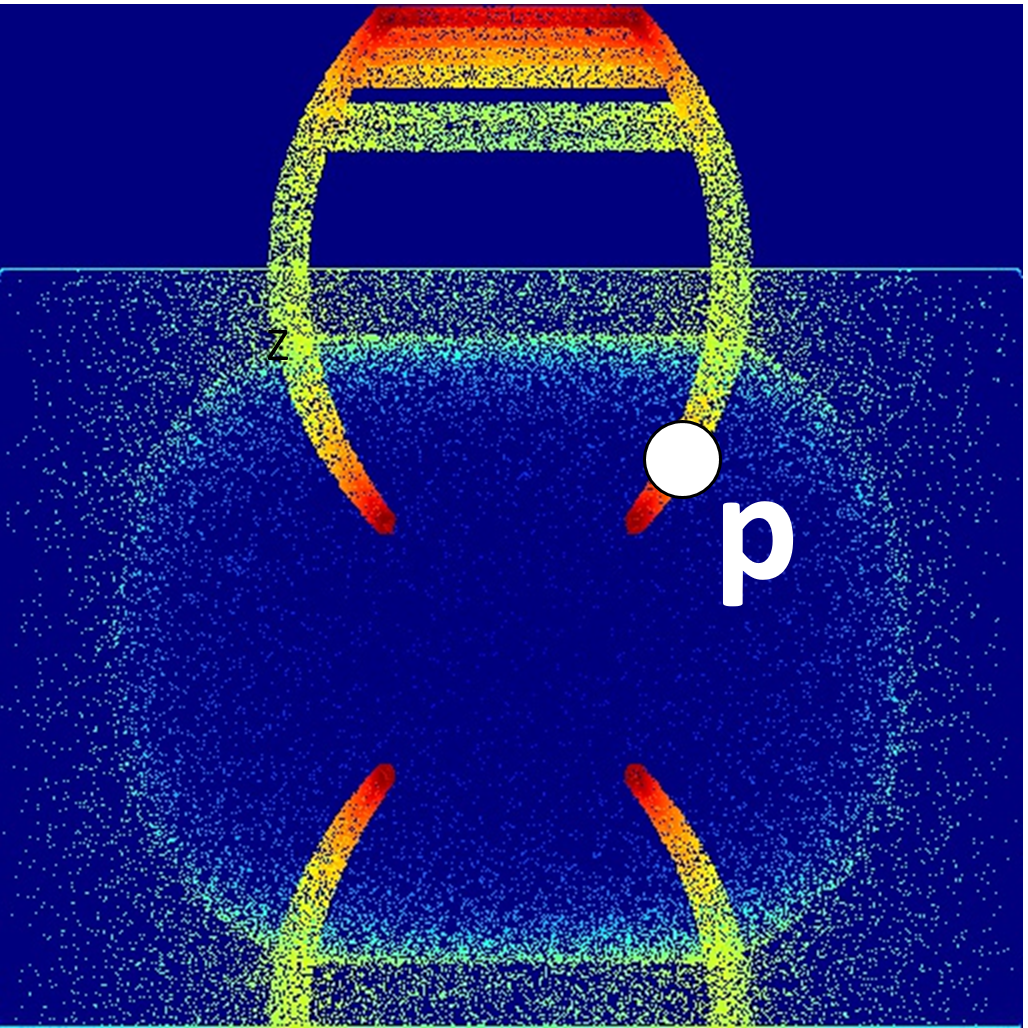}{\caption*{(d)\tiny Cassini}}}
\vfill
  {\label{fig:fig5}\includegraphics[width=\textwidth,height=1.5cm]{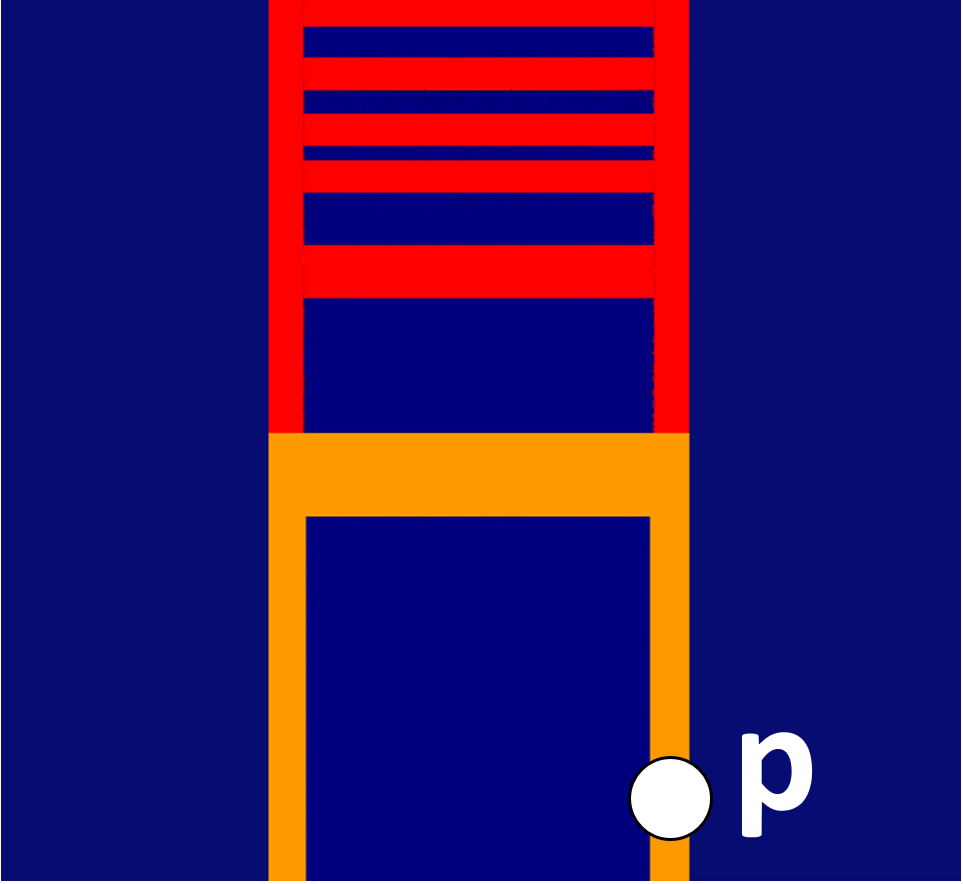}{\caption*{(g)\tiny Depth-map}}}
\end{minipage}
\begin{minipage}[b][\ht\measurebox][s]{0.15\textwidth}
\centering
  {\label{fig:fig6}\includegraphics[width=\textwidth,height=1.5cm]{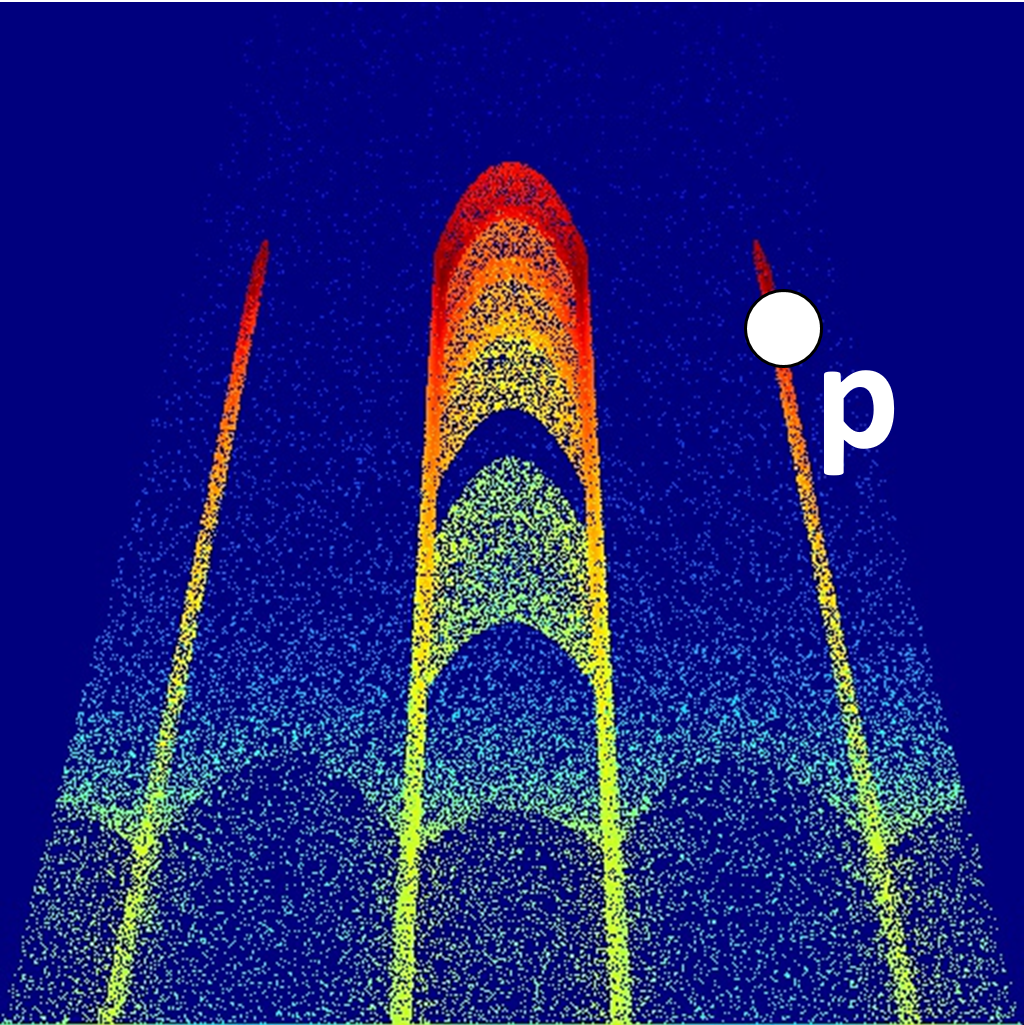}{\caption*{(e)\tiny Eckert IV}}}
\vfill
  {\label{fig:fig7}\includegraphics[width=\textwidth,height=1.5cm]{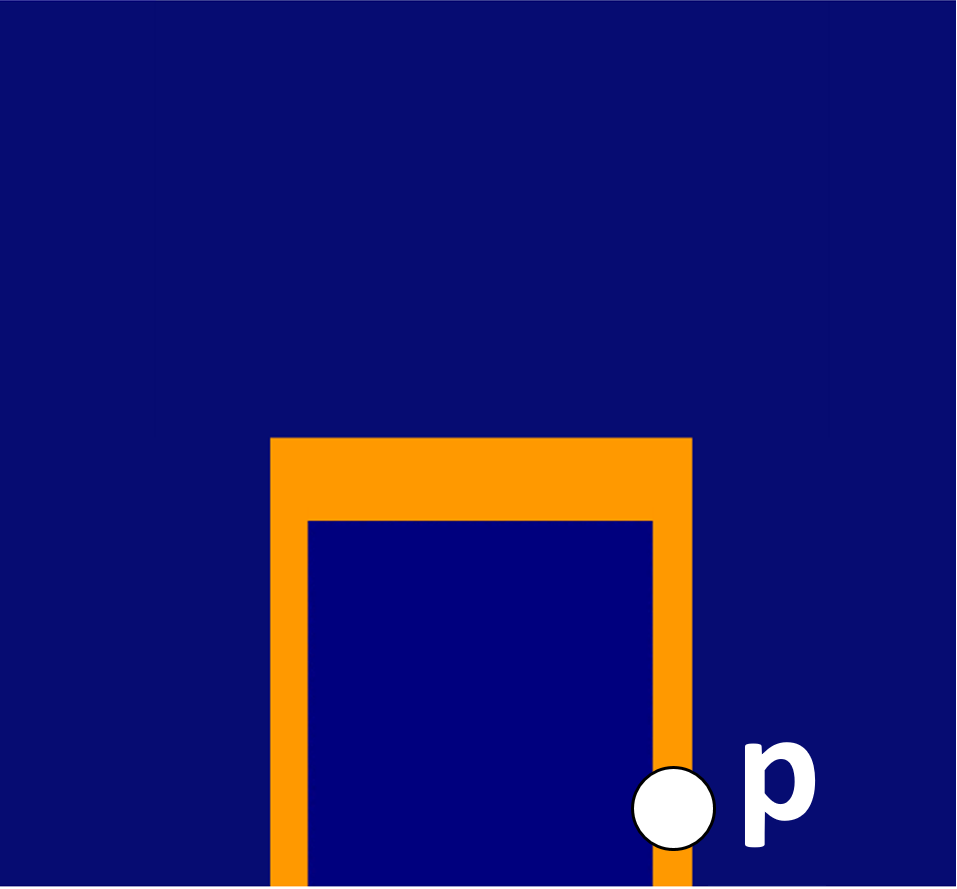}{\caption*{(h)\tiny Slice}}}
\end{minipage}
\caption{2D representation of surface of 3D object. (a) 3D mesh model with a point \(\mathit{p}\) at the surface and its corresponding unit vector \(\mathit{e}\) from the origin \(\mathit{0}\). (b)~(e) different types of stereographic projection functions. (f) Panoramic view \cite{7273863,Sfikas2017ExploitingTP,SFIKAS2018208}. (g) Depth-map \cite{7780912,7780783}. (h) Slice-based projection \cite{7965883}.}
\label{fig:streo}
\end{figure}

\noindent
\textbf{UV Projection \cite{nla.cat-vn2145201}:}
\begin{align}
    u &= 0.5 + \frac{\lambda}{2\pi},\\
    v &= 0.5 - \frac{\phi}{\pi},
\end{align}
\textbf{Kavrayskiy VII Projection \cite{nla.cat-vn2145201}:}
\begin{align}
    u &= \frac{3\lambda}{2}\sqrt{\frac{1}{3}-(\frac{\phi}{\pi})^2},\\
    v &= \phi,
\end{align}
\textbf{Eckert IV Projection \cite{nla.cat-vn2145201}:}
\begin{align}
    u &= 2\lambda\sqrt{\frac{4-3\sin|\phi|}{6\pi}},\\
    v &= \sqrt{\frac{2\pi}{3}}(2-\sqrt{4-3\sin|\phi|}),
\end{align}
\textbf{Cassini Projection \cite{nla.cat-vn2145201}:}
\begin{align}
    u &= 2\arcsin(\cos\phi\sin\lambda),\\
    v &= \arctan2(\frac{\tan\phi}{\cos\lambda}),
\end{align}

\noindent where, $\lambda=\arctan2{(e_x,e_y)}$ and $\phi=\arcsin{e_z}$ refer to the longitude and the latitude, respectively.

After determining the UV coordinates of the 3D object in the 2D mapped image, we set a value of each pixel with the distance of the corresponding point $\mathit{p}$ from the origin in the 3D object as shown in Fig.~\ref{fig:streo}(a). We discretize the 2D image to have a size of 128 x 128. As shown in Fig.~\ref{fig:streo}(b)-(h). We note that the stereographic representations of 3D object preserve more details about the shape of the 3D object compared to other approaches such as panorama  \cite{7273863,Sfikas2017ExploitingTP,SFIKAS2018208}, slice  \cite{7965883}, and multi-view \cite{7410471,kanezaki2018_rotationnet} representations.

\subsection{Network Architecture}
\begin{figure}
\centering
\includegraphics[height=3.5cm]{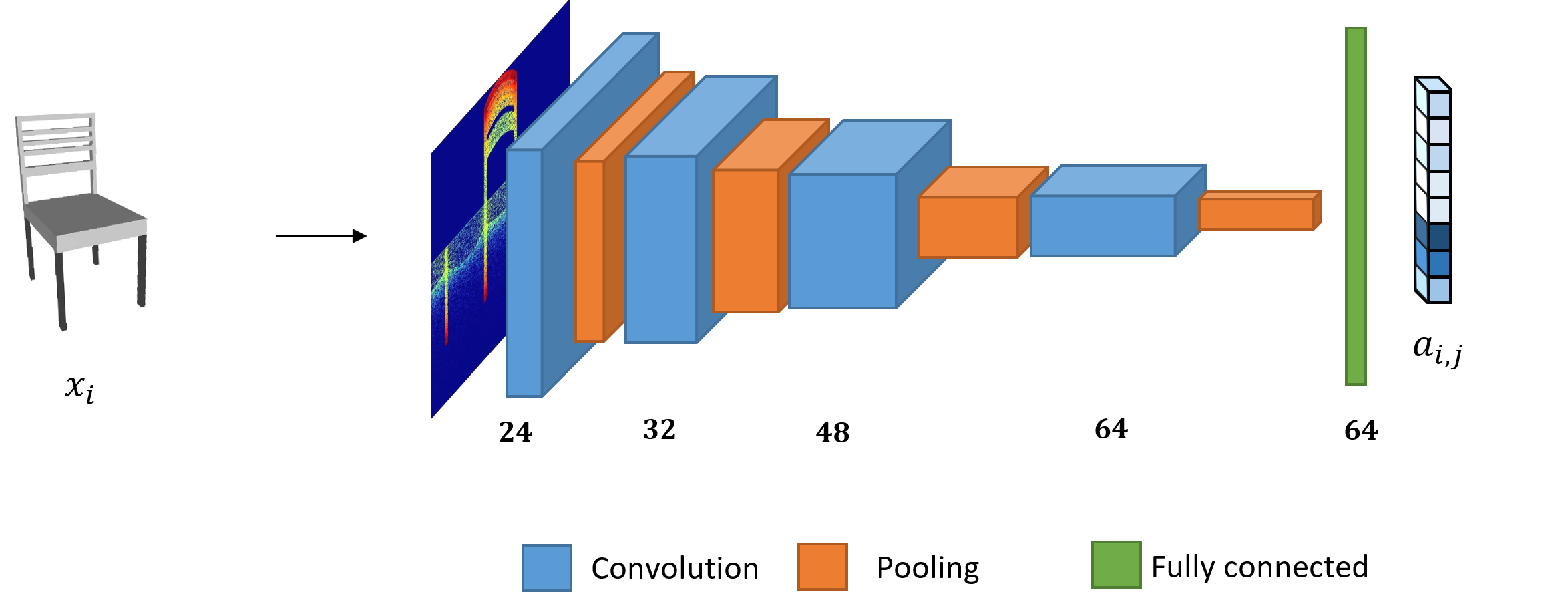}
\caption{Illustration of proposed SPNet, a shallow 2D convolutional neural network architecture. $a_{i,j}$ denotes the output from the last fully connected layer.}
\label{fig:network}
\end{figure}

We propose SPNet, a very shallow 2D CNN which consists of 4 convolutional layers and two fully connected layers. For each convolutional layer, we use a convolutional kernel of size 3x3 followed by tanh non-linearity and 2x2 max-pooling layers except for the last convolutional layer where we use global average pooling in place of max-pooling. Each side of inputs to all convolutional layers is zero-padded by 1 pixel to keep the feature map size unchanged. We also propose to add dropout after every layer except for the last fully connected layer to prevent over-fitting and for better generalization capability. The number of feature maps of our convolutional layers is 24, 32, 48, and 64, respectively. Details of the model are shown in Fig.~\ref{fig:network}.

\subsection{View Selection}
To construct multiple view stereographic representations from a 3D object, we augment the data with azimuth and elevation rotations. We first rotate the object along the gravity axis, each rotated $\ang{45}$ intervals. We further generate more views through elevation rotations with $\ang{45}$ intervals. Both angles are sampled uniformly from [0, $\ang{360}$] to generate $N=64$ views in total. Let us denote generated views of the object $x_i$ as $\{v_{i,j}\}_{j=1}^{N}$ where $\mathit{i}$ refers to the instance of the 3D object and $\mathit{j}$  refers to the rotated instance of the corresponding 3D object.

\begin{figure}[!b]
\centering
\includegraphics[width=11.5cm,height=4cm]{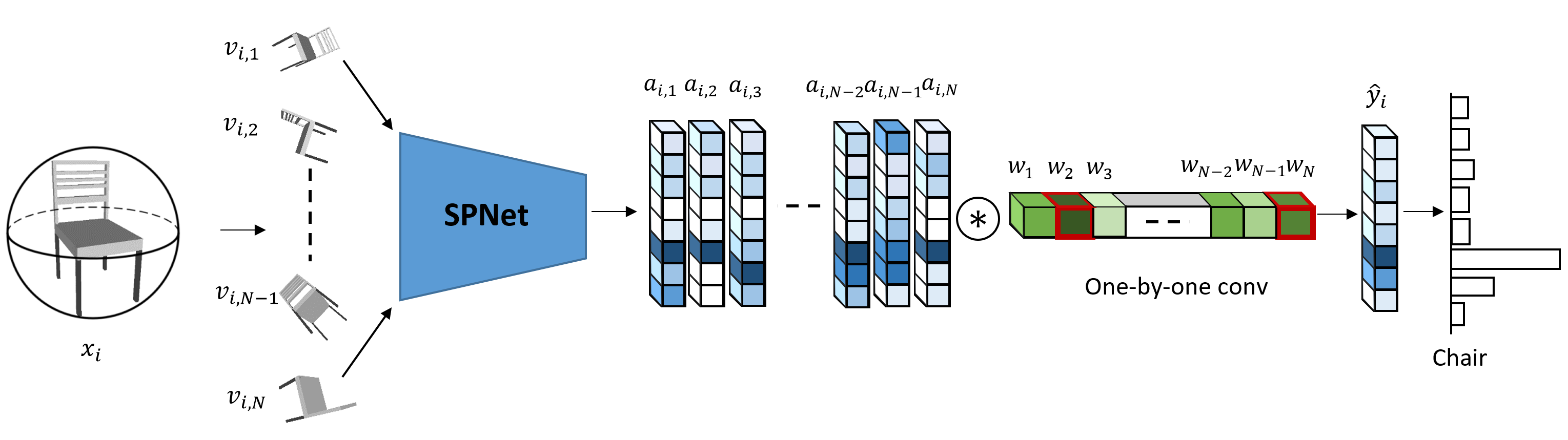}
\caption{Illustration of view selection and view ensemble. Both view selection and view ensemble adopt the same architecture but with different numbers of views to train each model. $a_{i,j}$ is the output of SPNet for the corresponding view $v_{i,j}$. Darker colors on the view-specific features $a_{i,j}$ and on the weights of the one-by-one convolutional layer denote higher values. Red boxes on the weights of the one-by-one convolutional kernel indicate the selected views.}
\label{fig:selection}
\end{figure}

All views $v_{i,j}$ are fed into the trained SPNet in Fig.~\ref{fig:network} to extract the view-specific feature response maps $a_{i,j}$. All  $\mathit{N}$ view-specific features are then passed through a one-by-one convolutional layer to perform weighted-average over all view-specific features. The output is then used as a final prediction score map. The overall process of view selection is visualized in Fig.~\ref{fig:selection}. The one-by-one convolutional layer in our view selection learns the importance of each view-specific features, thereby indicating the degree of contributions of each view to the final prediction. Once our view selection converges, we select $M$ most discriminative views $\{v^*_{i,j}\}_{j=1}^{M}$ where $M\leq N$ by observing the highest weight values in the one-by-one convolutional kernel.

\subsection{View Ensemble}

\begin{figure}
\centering
\includegraphics[width=12.2cm,height=3cm]{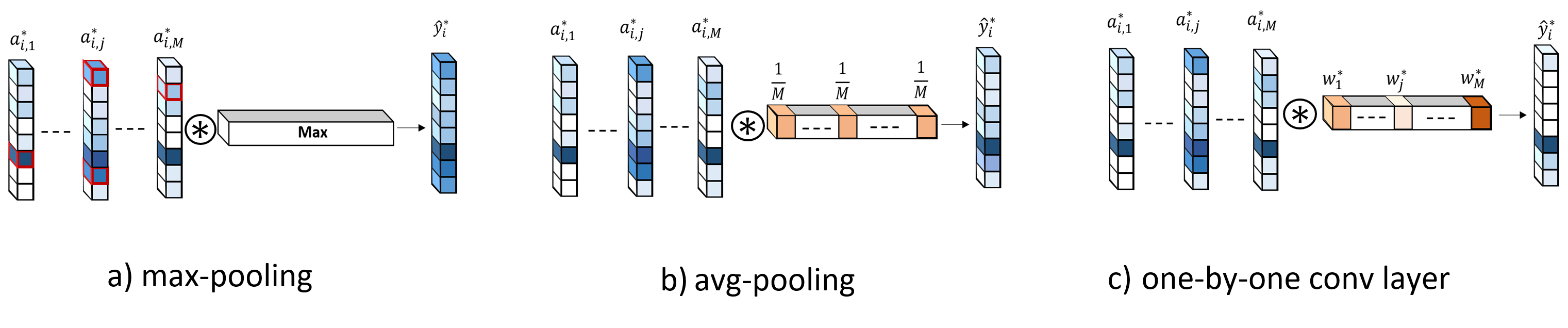}
\caption{Comparison of different types of ensemble. Darker colors on each view-specific features and weights of the one-by-one convolutional kernel indicate higher values.}
\label{fig:ensemble}
\end{figure}

Many recent works~\cite{Russakovsky2015,7540737,7769199} have shown that the use of ensemble technique provides a significant boost to the classification performance. Thus, we also exploit the weighted-average over predictions of M selected views $\{v^*_{i,j}\}_{j=1}^M$.

We train our view ensemble model in Fig.~\ref{fig:selection} by using only the selected most important M views $\{v^*_{i,j}\}_{j=1}^M$. Moreover, we examine different types of aggregation for the predictions of M selected views: 
\\
\textbf{Max-pooling}:
\begin{equation}
    \hat{y}^*_i = \max_{j}\{a^*_{i,j}\},
\end{equation}
\textbf{Avg-pooling}:
\begin{equation}
    \hat{y}^*_i = \frac{1}{M}\sum_{j=1}^{M} a^*_{i,j},
\end{equation}
\textbf{Weighted-average}:
\begin{equation}
    \hat{y}^*_i = \sum_{j=1}^{M} w^*_j a^*_{i,j},
\end{equation}

Where, $\hat{y}^*_i$ denotes the estimate of the object category label for each object $x_i$.

We have tested these three ensemble methods empirically and found that by learning the weights $\{w_j^*\}_{j=1}^{M}$ of the one-by-one convolutional layer properly, the weighted-average produces superior performance over the max-pooling and the average-pooling  \cite{7410471,7273863,Wang2017DominantSC,7780978,Wang2017OCNNOC,SFIKAS2018208}, as shown in Table~\ref{table:ensemble}.

\section{Experimental Evaluation}

\subsection{Datasets}
We have evaluated our method on the two subsets of the Princeton ModelNet large-scale 3D CAD model dataset  \cite{7298801} and the ShpeNet Core55, a subset of the ShapeNet dataset  \cite{Savva:2016:LSR:3056462.3056479}.

ModelNet-10 includes ten categories of 3991 and 908 models into training, and testing partitions, respectively. The dataset provides objects of same orientations. 

ModelNet-40 contains 12,311 CAD models split into 40 categories that provides objects of same orientations. The training and testing subsets consist of 9843 and 2468 models, respectively.

ShapeNet Core55 contains 51,300 3D models in 55 categories and several subcategories. Two versions of ShapeNet Core55 exist (a) consistently aligned 3D models and (b) models that are perturbed by random rotations. This dataset split into three subsets of 70\%, 10\% and 20\% for training, validation, and testing respectively. We trained and evaluated our 3D retrieval method on the training set and test set of the aligned 3D models, respectively.

\subsection{Training}

The baseline architecture of our CNN is shown in Fig.~\ref{fig:network} which is smaller than the VGG-M network architecture that MVCNN  \cite{7410471} used.
Table~\ref{table:models} shows the comparison of classification accuracy on the ModelNet-10  \cite{7298801} of our baseline architecture and some famous Convolutional Neural Network architectures.
To train SPNet, we used SGD optimizer with a learning rate of 0.01.

\begin{table}
\begin{center}
\caption{Classification accuracy on ModelNet-10 with various network architectures for a single view.}
\label{table:models}
\begin{tabular}{lccccccc}
\hline\noalign{\smallskip}
Architectures & SPNet (ours) & VGG-16 & ResNet-18 & ResNet-32 & ResNet-50 & ResNet-101\\
\noalign{\smallskip}
\hline
\noalign{\smallskip}
Accuracy & {\textbf{93.39\%}} & {83.92\%} & {91.74\%} & {91.19\%}  & {92.18\%} & {91.41\%}\\
\hline
\end{tabular}
\end{center}
\end{table}
\setlength{\tabcolsep}{1.4pt}

\subsection{Choice of Stereographic Projection}
We have evaluated several stereographic projection models for the 3D classification task including UV, Kavrayskiy VII, Eckert IV, and Cassini \cite{nla.cat-vn2145201}. 
Table~\ref{table:map} shows the test results on ModelNet-10 \cite{7298801}, where we can clearly observe that the UV-mapping outperforms the others. Since the UV-mapping is proven to be the best, we will use this mapping function in all subsequent experiments.

\begin{table}
\begin{center}
\caption{Classification accuracy on ModelNet-10 with various mapping functions.}
\label{table:map}
\begin{tabular}{lc}
\hline\noalign{\smallskip}
mapping function & accuracy\\
\noalign{\smallskip}
\hline
\noalign{\smallskip}
UV \cite{nla.cat-vn2145201} & {\textbf{93.39\%}}\\
Kavrayskiy VII \cite{nla.cat-vn2145201} & {93.17\%}\\
Eckert IV \cite{nla.cat-vn2145201} & {89.76\%}\\
Cassini \cite{nla.cat-vn2145201} & {92.51\%}\\
Depth-map (YZ-plane) & {85.02\%}\\
Panorama (around Z-axis)  & {92.07\%}\\
\hline
\end{tabular}
\end{center}
\end{table}
\setlength{\tabcolsep}{1.4pt}

\subsection{Test on View Selection Schemes}
We consider three view selection setups for the ensemble of the multi-view 2D stereographic representation to demonstrate the preference and power of our view selection approach.\\
\textbf{Case (i): Major axes} In this case, we set the viewpoints along three axes, x-axis, y-axis, and z-axis. The objects have same orientation namely that the viewpoint is along the x-axis. To obtain the two other viewpoints, each time we rotate the objects by
$\theta = \ang{90}$ 
and
$\phi = \ang{90}$ 
around z-axis and y-axis, respectively.\\
\textbf{Case (ii): 12 MVCNN} In this case, we fix z-axis as the rotation axis. We place the viewpoints at 
$\phi =\ang{30}$ 
from the ground plane and each time rotate the objects by 
$\theta=\ang{30}$
around the z-axis to obtain 12 views for the object.\\
\textbf{Case (iii): View Selection} Our view selection method which learns the view's influence by a one-by-one convolutional layer. We used the method on 64 different rotations by rotating the objects around z-axis and y-axis and then selected the views with the highest influence.

We compared the classification accuracy for these three view setup on the ModelNet-10  \cite{7298801} with our view ensemble neural network architecture named SPNet\_VE. Table~\ref{table:ensemble} shows the comparison of classification accuracy on the ModelNet-10  \cite{7298801} of plain and ensemble with the Max-pooling, Avg-pooling, and one-by-one convolutional layer as a weighted-average over the score features of the multi-view 2D representations. From these results, we observe that our learned weighted averaging of 5 views gives the best performance over other schemes, so that we use this ensemble model for our experiments.

\begin{table}[t]
\begin{center}
\caption{Classification accuracy on ModelNet-10 with various view selection schemes.}
\label{table:ensemble}
\begin{tabular}{lcccc}
\hline\noalign{\smallskip}
View setup & \#views & Max-pool & Avg-pool & one-by-one conv\\
\noalign{\smallskip}
\hline
\noalign{\smallskip}
Plain & 1  & {93.39\%} & {93.39\%}& {93.39\%}\\
\hline
Major axes & 3 & {95.15\%} & {95.59\%}& {96.26\%}\\
\hline
MVCNN & 12 & {91.63\%} & {92.51\%}& {92.40\%}\\
\hline
\multirow{7}{*}{View Selection }
& 1 & {93.39\%} & {93.39\%}& {93.39\%}\\
& 2 & {95.82\%} & {96.15\%}& {96.15\%}\\
& 3 & {95.59\%} & {95.59\%}& {96.26\%}\\
& 4 & {95.15\%} & {95.48\%}& {96.58\%}\\
& 5 & {94.05\%} & {95.93\%}& {\textbf{97.25}\%}\\
& 6  & {94.16\%} & {95.15\%}& {97.03\%}\\
& 64  & {90.64\%} & {91.74\%}& {91.52\%}\\
\hline
\end{tabular}
\end{center}
\end{table}
\setlength{\tabcolsep}{1.4pt}

\subsection{3D Object Classification}
We have first evaluated our baseline method SPNet in classification on both ModelNet-10 \cite{7298801} and ModelNet-40 \cite{7298801}. The performance of our model is measured by the average binary categorical accuracy.

We have compared our method with recent sate-of-the-art methods including 3D ShapeNet  \cite{7298801}, GIFT  \cite{7780912}, DeepPano  \cite{7273863}, Multi-view Convolutional Neural Networks (MVCNN)  \cite{7410471}, Geometry Image descriptor  \cite{Sinha2016DeepL3}. In addition to above methods the results are extended to include  the following voxel based methods: ORION  \cite{Sedaghat2016OrientationboostedVN}, 3D-GAN  \cite{Wu:2016:LPL:3157096.3157106}, VoxNet  \cite{7353481}, O-CNN  \cite{Wang2017OCNNOC} and OctNet  \cite{8100184}.
Table~\ref{table:class} summarizes the comparative results of classification on ModelNet-10 and ModelNet-40 in terms of  GPU memory usage and the number of parameters during the training phase, and classification accuracy.

We note that in our approach, the view-ensemble model (SPNet\_VE) boosts significant performance improvement over the baseline model (SPNet) by $3.9\%$ and $4.0\%$ on ModelNet-10 and ModelNet-40, respectively.
Moreover, SPNet\_VE achieved comparable results to those of the state-of-the-arts RotationNet \cite{kanezaki2018_rotationnet}, while requiring much less memory ($2\%$) and network parameters ($0.2\%$), respectively. 
Note also that there is a large gap between the average $(94.82\%)$ and maximum $(98.46\%)$ accuracy of the RotationNet \cite{kanezaki2018_rotationnet} which shows this method is not stable while our method showed consistent performances $(97.25\%)$ for each trial of training process.

\begin{table}[t]
\begin{center}
\caption{Classification results and comparison to state-of-the-art methods on ModelNet-10 and ModelNet-40. Also the number of parameters and GPU memory usage. VE indicates view ensemble.}
\label{table:class}
\begin{tabular}{clcccc}
\noalign{\smallskip}\hline
\centering
InputModality &
Method &
{GPU memory}   &
{Parameters} &
\multicolumn{2}{l}{ModelNet}  \\
 \cline{5-6}
\noalign{\smallskip}
 & & &  & class 10 & class 40\\
\noalign{\smallskip}
\hline
\noalign{\smallskip}
\multirow{2}{*}{Point Clouds}
& PointNet \cite{8099499}& {-}  & {3.5M} & {-} & {89.2\%}\\
& PointNet++ \cite{qi2017pointnetplusplus}& {-}  & {-} & {-} & {91.9\%}\\
\hline
\multirow{10}{*}{3D Volume}
& ShapeNet  \cite{7298801}  & {60.5MB}  & {15M} & {83.50\%} & {77.00\%}\\
& LightNet \cite{ZHI2018199}  & {2MB}  & {0.3M} & {93.39\%} & {86.90\%}\\
& ORION  \cite{Sedaghat2016OrientationboostedVN}  & {4.5MB}  & {0.91M} & {93.80\%} & {-}\\
& VRN \cite{Brock2016GenerativeAD}  & {129MB}  & {18M} & {93.60\%} & {91.33\%}\\
& VRN Ensemble  \cite{Brock2016GenerativeAD} & {678MB} & {93.5M} & {97.14\%} & {95.54\%}\\
& VoxNet \cite{7353481} & {4.5MB}  & {0.9M} & {92.00\%} & {83.00\%}\\
& FusionNet \cite{ZHI2018199} & {548MB}  & {118M} & {93.10\%} & {90.80\%}\\
& 3D-GAN  \cite{Wu:2016:LPL:3157096.3157106}  &{56MB}  & {11M} & {91.00\%} & {83.30\%}\\
& OctNet \cite{8100184}  & {-}  & {-} & {90.42\%} & {-}\\
& O-CNN  \cite{Wang2017OCNNOC}  & {-}  & {-} & {-} & {90.6\%}\\
\hline
\multirow{2}{*}{Others}
& Spherical CNNs  \cite{s.2018spherical} & {-} & {1.4M}  & {-} & {-}\\
& LonchaNet  \cite{7965883} & {-}  & {15M} & {94.37\%} & {-}\\
\hline
\multirow{3}{*}{2D Represen.}
& MVCNN   \cite{7410471} & {331MB}  & {42M} & {-} & {90.10\%}\\
& MVCNN-MultiRes  \cite{7780978} & {-}  & {180M} & {-} & {91.40\%}\\
& RotationNet \cite{kanezaki2018_rotationnet} & {731MB}  & {42M} & {\textbf{98.46\%}} & {\textbf{97.37\%}}\\
\hline
\multirow{7}{*}{2.5D Represen. }
& DeepPano  \cite{7273863} & {9.8MB}  & {3.27M} & {85.45\%} & {77.63\%}\\
& PANORAMA-NN \cite{Sfikas2017ExploitingTP} & {6.77MB}  & {2.86M} & {91.10\%} & {90.70\%}\\
& PANORAMA-ENN   \cite{SFIKAS2018208} & {42MB}  & {8.6M} &  {96.85\%} & {95.56\%}\\
& GIFT  \cite{7780912} & {-}  & {-} &  {92.35\%} & {83.10\%}\\
& Pairwise  \cite{7780783} & {-}  & {42M} & {92.80\%} & {90.70\%}\\
& SPNet (ours) & {3MB}  & {86K}  & {93.39\%} & {88.61\%}\\
& SPNet\_VE (ours) & {15MB}  & {86K}  & {97.25\%} & {92.63\%}\\
\hline
\end{tabular}
\end{center}
\end{table}

\subsection{Shape Retrieval}

We have evaluated the view ensemble version, SPNet\_VE with the learned five views for the 3D object retrieval task under three datasets, ModeNet-10  \cite{7298801}, ModelNet-40  \cite{7298801} and ShepeNet Core 55  \cite{Savva:2016:LSR:3056462.3056479}.
Table~\ref{table:retrive} shows the results of our retrieval experiment on the test sets of ModelNet-10 and ModelNet-40 with mean Average Precision (mAP) in comparison with other state-of-the-art methods.

We used the learned global features of our ensemble network before the last $\tanh$ activation function. Then, we applied the softmax function to create the best feature descriptors for all 3D objects. We sorted the most relevant 3D objects for each query from the test set by using both 
$L_1$
and
$L_2$
distance metrics. Our SPNet\_VE with 
$L_1$ 
achieved the best performance on ModelNet-10 and the second best on ModelNet-40. Note that the complexity of our model is much lighter than PANORAMA-ENN \cite{SFIKAS2018208}; only $36\%$ and $1\%$ of the memory and parameters of PANORAMA-ENN are used, respectively. 

\begin{table}[!t]
\begin{center}
\caption{Comparison of retrieval results measured in mean Average Precision (mAP) on the ModelNet-10 and ModelNet-40 datasets.}
\label{table:retrive}
\begin{tabular}{lcccc}
\noalign{\smallskip}\hline
\centering
Method &
{GPU memory}   &
{Parameters} &
\multicolumn{2}{l}{ModelNet(mAP)}  \\
 \cline{4-5}
\noalign{\smallskip}
 & &  & class 10 & class 40\\
\noalign{\smallskip}
\hline
\noalign{\smallskip}
MVCNN \cite{7410471}  & {331MB} & {42M} & {-} & {79.5\%}\\
Geometry Image \cite{Sinha2016DeepL3} & {-} & {-} & {74.9\%}  & {51.3\%}\\
GIFT \cite{7780912} & {-} & {-} &  {91.12\%}  & {81.94\%}\\
DeepPano \cite{7273863} & {9.8MB} & {3.27M} & {84.18\%}  & {76.81\%}\\
3D ShapeNets \cite{7298801} & {-} & {-} & {68.3\%}  & {49.2\%}\\
PANORAMA-ENN   \cite{SFIKAS2018208} & {42MB} & {8.6M} & {93.28\%} & {\textbf{86.34\%}}\\
SPNet\_VE (L2)  & {15MB} & {86K} & {92.94\%} & {84.68\%}\\
SPNet\_VE (L1)  & {15MB} & {86K} & {\textbf{94.20\%}} & {85.21\%}\\
\hline
\end{tabular}
\end{center}
\end{table}

\begin{figure}[t]
\centering
\includegraphics[width=11cm,height=8cm]{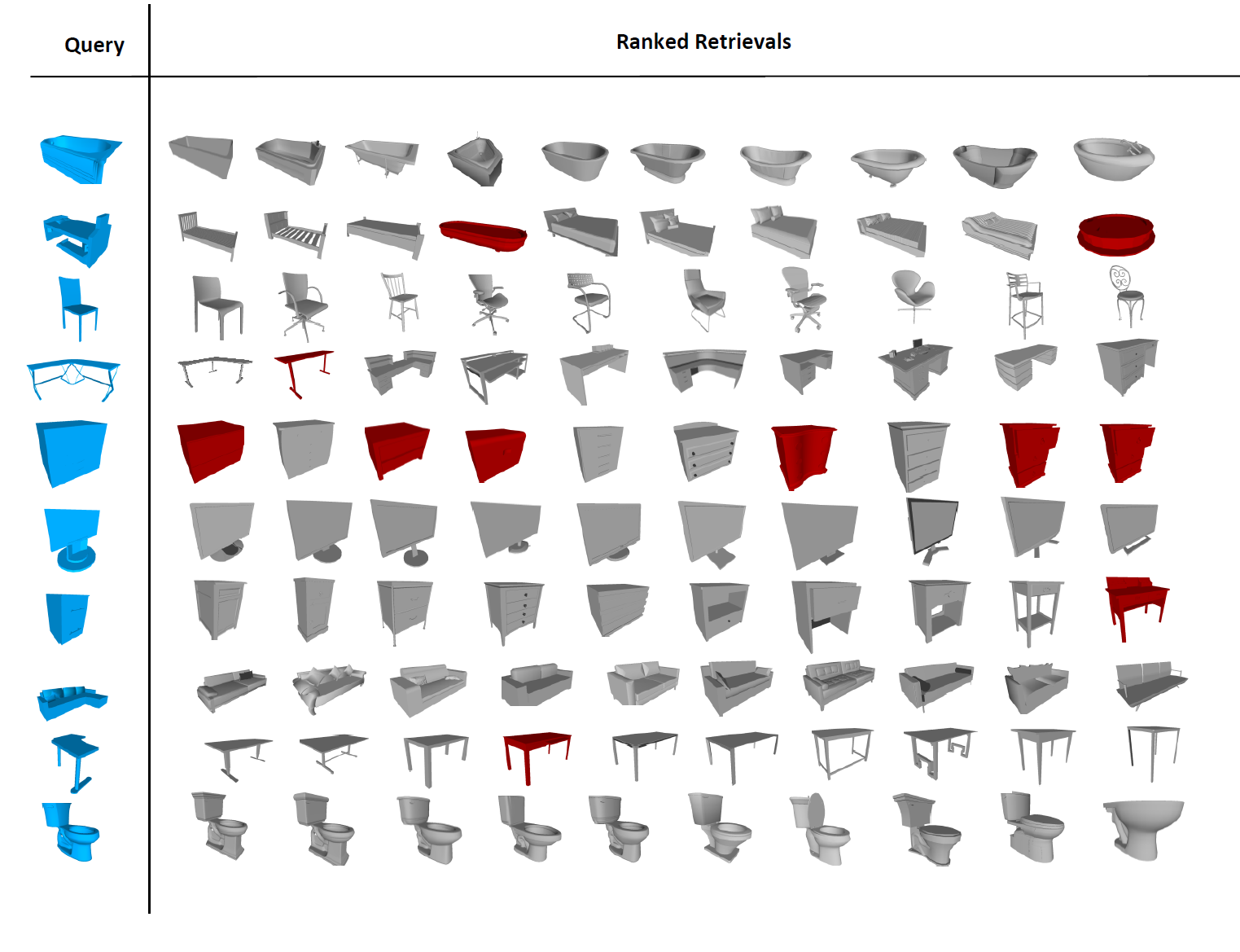}
\caption{Retrieval examples by the proposed SPNet\_VE on the test set of the ModelNet-10 dataset. The first column illustrates the queries and the remaining columns show the corresponding retrieved models in rank order. Retrieved objects with blue and red colors are queries and failure cases, respectively.}
\label{fig:retrievals}
\end{figure}

\begin{table}[!b]
\begin{center}
\caption{Retrieval results measured in F-score, mean Average Precision (mAP) and Normalized Discounted Gain (NDCG) on the normalized ShapeNet Core55. VE indicates View Ensemble.}
\label{table:retrieve55}
\begin{tabular}{lccccccc}
\noalign{\smallskip}\hline
Method &
\multicolumn{3}{c}{Micro-averaged}  & &
\multicolumn{3}{c}{Macro-averaged}   \\
\cline{2-4}\cline{6-8}
\noalign{\smallskip}
 & F-score & mAP & NDCG & & F-score & mAP & NDCG \\
\noalign{\smallskip}
\hline
\noalign{\smallskip}
Kanezaki & {\textbf{79.8\%}}  & {\textbf{77.2\%}} & {86.5\%}& & {\textbf{59.0\%}} & {\textbf{58.3\%}} & {65.6\%}\\
Zhou & {76.7\%}  & {72.2\%} & {82.7\%}& & {58.1\%} & {57.5\%} & {65.7\%}\\
Tatsuma & {77.2\%}  & {74.9\%} & {82.8\%}& & {51.9\%} & {49.6\%} & {55.9\%}\\
FUruya & {71.2\%}  & {66.3\%} & {76.2\%}& & {50.5\%} & {47.7\%} & {56.3\%}\\
Thermos & {69.2\%}  & {62.2\%} & {73.2\%}& & {48.4\%} & {41.8\%} & {50.2\%}\\
Deng & {47.9\%}  & {54.0\%} & {65.4\%}& & {16.6\%} & {33.9\%} & {40.4\%}\\
Li & {28.2\%}  & {19.9\%} & {33.0\%}& & {19.7\%} & {25.5\%} & {37.7\%}\\
Mk & {25.3\%}  & {19.2\%} & {27.7\%}& & {25.8\%} & {23.2\%} & {33.7\%}\\
SHREC16-Su & {76.4\%}  & {73.5\%} & {81.5\%}& & {57.5\%} & {56.6\%} & {64.0\%}\\
SHREC16-Bai & {68.9\%}  & {64.0\%} & {76.5\%}& & {45.4\%} & {44.7\%} & {54.8\%}\\
SPNet\_VE & {78.9\%} & {69.2\%} & {\textbf{89.0\%}} & & {53.5\%} & {39.2\%} & {\textbf{69.5\%}}\\
\hline
\end{tabular}
\end{center}
\end{table}
\setlength{\tabcolsep}{1.4pt}

Table~\ref{table:retrieve55} shows our results of the retrieval experiment on the large-scale normalized ShapeNet Core55 dataset. We tested our ensemble model by F-score, mean Average Precision (mAP) and Normalized Discounted Gain (NDCG) metrics in comparison to  \cite{7780912,Furuya2016DeepAO}. The Macro-averaged is an unweighted average over the entire dataset while the Micro-averaged gives an average over category. The proposed method outperformed the other methods by NDCG metric on both the Macro and Micro averaged.

\begin{figure}[!b]
\centering
\includegraphics[height=6.0cm]{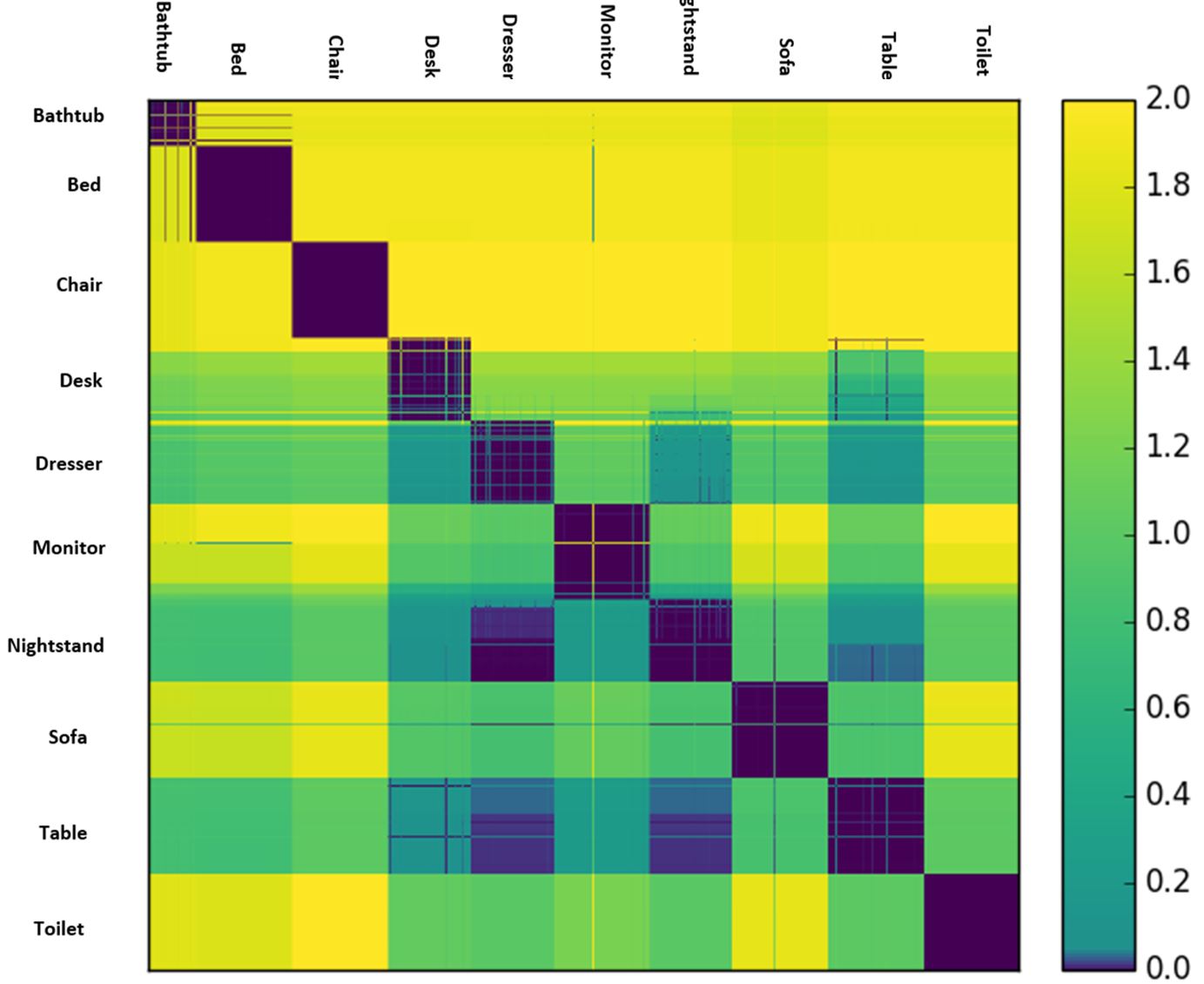}
\caption{Confusion matrix for 3D objects on the test set of the ModelNet-10. Values of the matrix show the similarity between pairs of 3D objects. Higher values indicate the two 3D objects have fewer similarities; see the color bar. }
\label{fig:confusion}
\end{figure}

Fig.~\ref{fig:retrievals} shows some of the retrieval cases on the test set of the ModelNet-10. The first column in the figure illustrates the queries and the remaining columns illustrate the corresponding retrieved objects in rank order. The red models indicate that the retrieved objects are in a wrong class with the queries. In other cases, the queries and the retrieved objects have the same classes. For instance, in the class of the dresser, the retrieved objects are so similar to the query while they are from different classes. The reason for these failure cases is that some objects from two different classes are hard to distinguish. Note that our approach does not have any failure cases in the class of Chair and Toilet of the ModelNet-10. 
Fig.~\ref{fig:confusion} shows the confusion matrix for all 3D objects on the test set of ModelNet-10. The similarity is measured by L1 distance. Therefore, so lower values indicate higher similarities between pairs of objects.

\subsection{Implementation}
We have evaluated the proposed method SPNet on an Intel (R) Core (TM) i5 @ 3.4GHz CPU system, with 32GB RAM and NVIDIA (R) GTX 1080 Ti GPU with 12GB RAM. The system was developed in Python 3.5.2, and the network was implemented using TensorFlow-1.4.0 via CUDA instruction set on the GPU. The runtime of our SPNet and the prepossessing per each object are 2.5ms and 120ms, respectively.

\section{Conclusions}
We proposed a novel ensemble architecture to learn 3D object descriptors based on the Convolutional Neural Networks. We used stereographic transformation to project 3D objects into a 2D planar followed by 2D CNNs to give confidence scores for multiple views. A one-by-one convolutional layer learns the importance of each view and selects the best views ordinary. To improve the performance, we proposed an ensemble CNN which combines the responses from the chosen views by weighted-averaging with learned weights. We evaluated our network on two large-scale datasets, ModelNet, and ShapeNet Core55. We showed that the performance of the proposed method for the classification task is par to those of the state-of-the-art approaches, while outperforms most existing works in the retrieval task. Moreover, our proposed model is most efficient regarding GPU memory usage and the number of parameters compared to existing networks. 

In the future works, the ensemble neural network can be extended. Moreover, The datasets that we used do not contain texture and color information. The one channel 2D plane represented by our stereographic representation could be extended to more channels if this information existed.

\section*{Acknowledgment}
This work was supported by the National Research Foundation of Korea (NRF) grant funded by the Korea Government (MSIT) (No.  NRF-2017R1A2B2011862).

%

\end{document}